\documentclass[letterpaper]{article}
\usepackage[labelsep=period]{caption} 
\usepackage{interspeech2010,amssymb,amsmath,epsfig}
\usepackage[safe]{tipa}
\usepackage{tipx,nicefrac,url}
\usepackage{fixltx2e} 
\usepackage{microtype} 
\usepackage{hyperref}
\makeatletter
\newcommand{\doitext}{doi:}
\newcommand*{\doi}{%
  \begingroup 
  \lccode`\~=`\#\relax 
  \lowercase{\def~{\#}}%
  \lccode`\~=`\_\relax
  \lowercase{\def~{\_}}%
  \lccode`\~=`\<\relax 
  \lowercase{\def~{\textless}}%
  \lccode`\~=`\>\relax 
  \lowercase{\def~{\textgreater}}%
  \lccode`\~=0\relax 
  \catcode`\#=\active 
  \catcode`\_=\active 
  \catcode`\<=\active 
  \catcode`\>=\active 
  \@doi
}
\def\@doi#1{%
  \let\#\relax
  \let\_\relax
  \let\textless\relax 
  \let\textgreater\relax 
  \edef\x{\toks0={{#1}}}%
  \x
  \edef\#{\@percentchar23}%
  \edef\_{_}%
  \edef\textless{\@percentchar3C}
  \edef\textgreater{\@percentchar3E}
  \edef\x{\toks1={\noexpand\href{http://dx.doi.org/#1}}}%
  \x
 \edef\x{\endgroup\the\toks1 {\doitext\the\toks0}}%
  \x
}
\makeatother

\makeatletter
\def\url@leostyle{%
  \@ifundefined{selectfont}{\def\UrlFont{\sf}}{\def\UrlFont{\small\ttfamily}}}
  \makeatother
\usepackage{listings}
\usepackage{color}
\hypersetup{ colorlinks=true, citecolor=blue, linkcolor=blue, urlcolor=blue }

\def\bysame{\leavevmode\hbox to3em{\hrulefill}\thinspace} 

\title{
\vspace{-5mm}
Motion trails from time-lapse video
\vspace{-4mm}
}

\makeatletter
\def\name#1{\gdef\@name{#1\\}}
\makeatother

\name{\vspace{-4mm}Camille Goudeseune}
\address{University of Illinois at Urbana-Champaign \\
{\small \tt cog@illinois.edu}}

\begin{document}
\maketitle
\begin{abstract}

From an image sequence captured by a stationary camera,
background subtraction can detect moving foreground objects in the scene.
Distinguishing foreground from background is further improved by various heuristics.
Then each object's motion can be emphasized by duplicating its positions as a motion trail.
These trails clarify the objects' spatial relationships.
Also, adding motion trails to a video before previewing it at high speed
reduces the risk of overlooking transient events.

\end{abstract}
\section{Motivation}

As cameras are becoming more common, so too are sequences of images captured by
stationary cameras, as either time-lapse photography or video.
Because of cameras' inexpensive data storage and long battery life,
such image sequences are commonly too long to preview in real time.
Unfortunately, faster previewing risks overlooking interesting brief transient events.
This paper presents a way to greatly reduce such risk:
video processing that increases the saliency of transient events, anomalies, and patterns,
letting them be seen at a glance during fast-forward or even in still frames,
instead of taxing short-term memory.
The approach is inspired by the solution to a similar problem:
in long audio recordings, increasing the saliency of brief anomalous sounds~\cite{timeliner}.

\begin{figure}[bp]
\centerline{\epsfig{figure=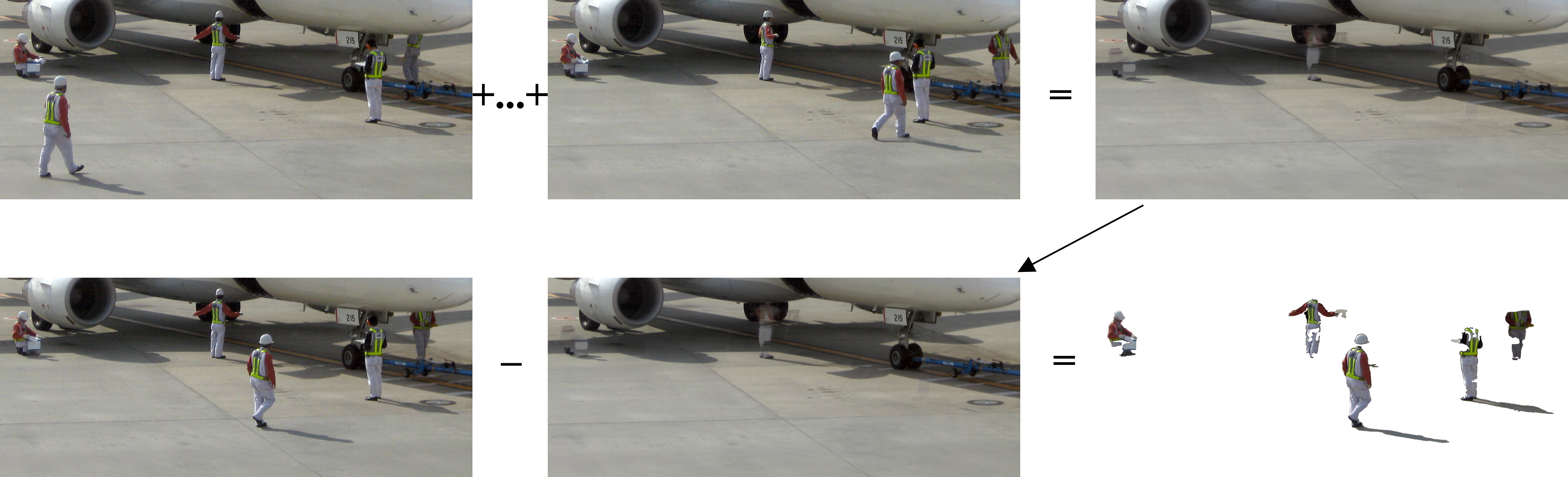, width=\linewidth}}
\caption{Background subtraction.
Top: averaging a sequence of images extracts their background.
Bottom: subtracting that background from another image extracts its foreground.}
\label{fig:fgbg}
\end{figure}

\begin{figure*}[htp] 
\centerline{\epsfig{figure=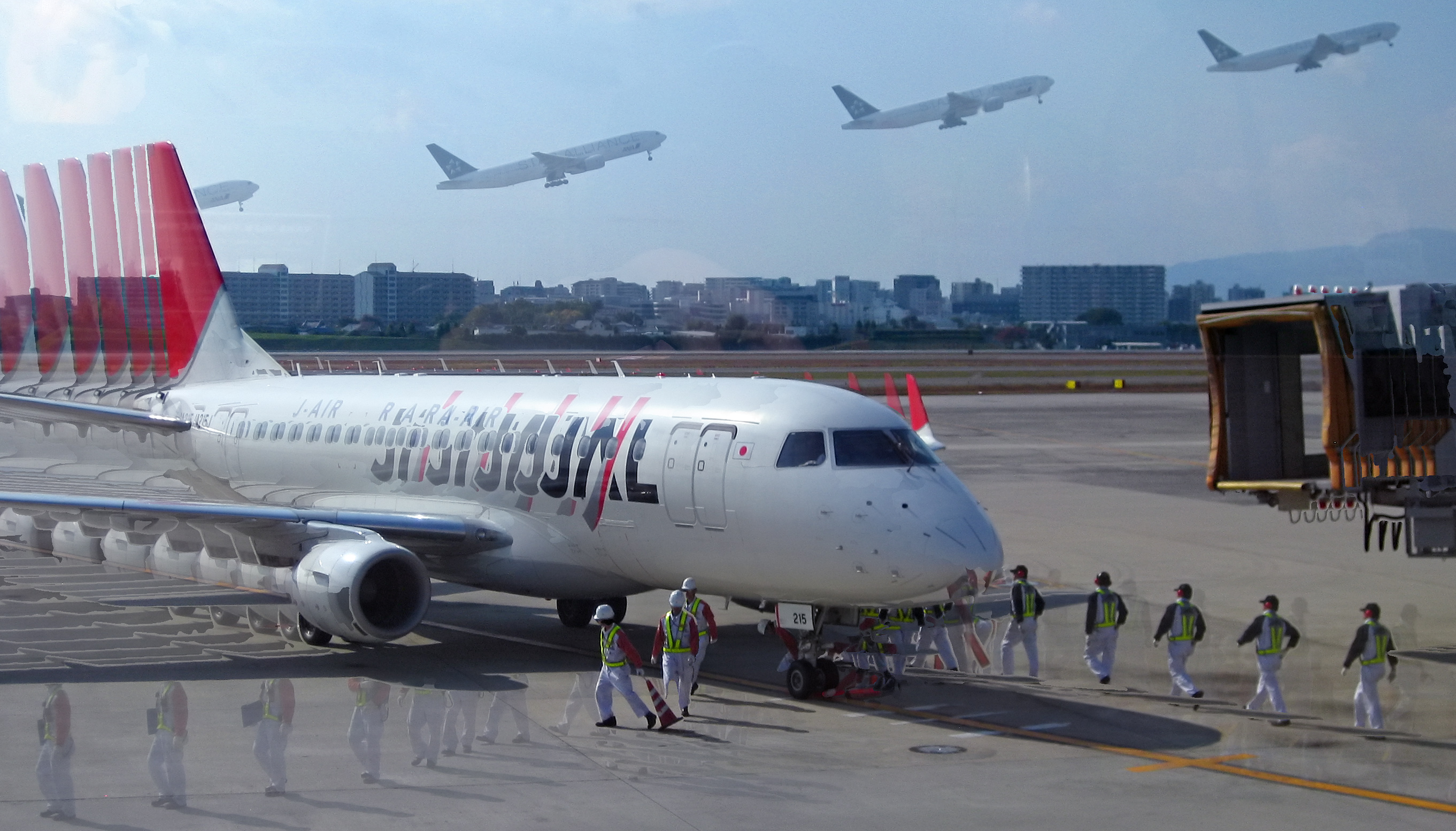, width=\linewidth}}
\caption{Motion trails of ground crew, airplanes, and an extending jet bridge.  Osaka International Airport, 2012-11-09.}
\label{fig:butterfly}
\end{figure*}

Given such a sequence of images, then, for each image we average
its neighboring images to smooth away any moving foreground objects.
What remains is the background part of each image.
Subtracting each background image from its corresponding original image
then yields a foreground image (fig.~\ref{fig:fgbg}).
Onto each background we then overlay multiple foregrounds,
thereby adding a motion trail to each object (fig.~\ref{fig:butterfly}).

Motion trails made with software such as Adobe's After Effects require
the foreground and background to be already separated into different
`layers.'  In fact, the foreground is often text rather than something from
a natural scene.  Such software implements only section~\ref{sec:trails}
of this paper.
Motion trails can also be made by replacing each video frame
with an average of its neighbors (section~\ref{sec:bg} by itself).
But then all foregrounds and backgrounds blur together:
every foreground must be as faint as those at the bottom left of
fig.~\ref{fig:butterfly}.
Finally, for the special case of nighttime, moving lights can be turned into bright streaks
by simply having each pixel position accumulate the brightest value encountered so far.
However, because these streaks cannot fade, they eventually drown out new lights,
making this shortcut impractical for long image sequences.

\begin{figure*}[htp]
\centerline{\epsfig{figure=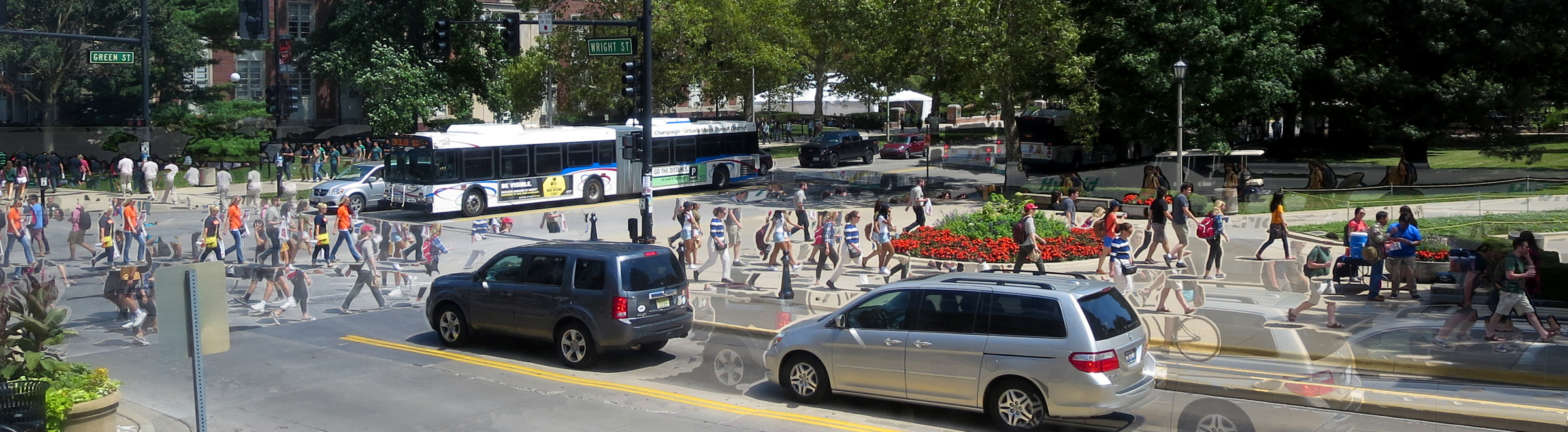, width=\linewidth}}
\caption{Mutually interfering motion trails.
Urbana-Champaign campus, 2015-08-21.}
\label{fig:wrightgreen}
\end{figure*}

This novel video processing is best suited to foreground objects that are small and sparse:
adding trails to a dense crowd of pedestrians merely adds clutter (fig.~\ref{fig:wrightgreen}).
Also, for trails to be visible at all,
images should be captured frequently enough to record multiple points along an object's path:
a construction site photographed only once per minute makes
workers and trucks appear and vanish, as if teleported.

Note that the images shown here are only excerpts from
the videos actually produced by the software described herein.

\section{Eliminating camera shake}
\label{sec:deshake}

A camera may move slightly with respect to its viewed scene
when its support experiences vibration, jostling, wind gusts, or even thermal expansion.
This camera movement may be visible as a shift from one image to the next,
especially if the image resolution exceeds 5 megapixels or if the lens has a long focal length.
Averaging the images in an interval that includes such movement
would undesirably combine backgrounds into a double exposure.
Fortunately, such camera motion can be suppressed with a translation in the image plane of a few pixels.
(A full homography is not needed in practice,
because the camera's translation is negligible,
and its rotation is only about the axes through its sensor plane.)
After accumulating these consecutive frame-to-frame translations,
each unshaken image is then cropped by the appropriate number of pixels,
to restore a common rectangular image field to the entire sequence.
We now consider the deshaking algorithm in detail.

First, the translational offset from each image to its successor is measured
by brute force correlation.
Both images are converted to grayscale.
Then one image is offset by $(x,y)$ for all $|x|$ and $|y|$ less than
some threshold.  (Run time is quadratic in that threshold.)  Whichever
offset minimizes the RMS pixelwise difference with the other image
is accepted as best.\footnote{
This offset can be refined to subpixel accuracy, by fitting a paraboloid
to nearby offsets and choosing the paraboloid's apex.
But such computation is rarely warranted,
because the images are usually downsampled before being finally rendered as a movie.
At any rate, efficient and intricate subpixel algorithms exist~\cite{subpixel}.}

Actually, the algorithm is less na\"ive.
Instead of finding the best offset for the whole image,
it does so for each of many square blocks within the image.
Each block is a few times larger than the largest expected offset,
typically a few dozen pixels.\footnote{
In practice, block size only slightly affects robustness against foreground motion.}
Low-contrast blocks such as blue sky or smooth pavement are ignored,
lest their spurious and noisy offsets corrupt those from more trustworthy blocks.
Once these offsets are found, an estimate is made of their geometric median,
which robustly ignores outliers, that is, moving objects.
This median is thus more accurate than the ``mean'' given by the best whole-image offset.

Incrementally summing these successive offsets yields each image's cumulative offset.
This in turn yields how much each image must be cropped to restore a common field of view to the entire sequence.
(For example, if some image is offset to the left by 5~pixels,
then some other image's right edge must also be cropped by 5~pixels.)
Finally, each image is offset and cropped, producing a new image sequence with camera motion eliminated.

Unfortunately, even the geometric median can be fooled if enough similar nonzero offsets occur.
For example, in a scene of mostly windblown clouds with a sliver of immobile prairie,
the cloudscape overwhelms the landscape, tricking the algorithm into reporting camera motion.
Because this scene is the same as stationary clouds with a sliver of dashboard seen from a fast car,
or fast clouds from a stopped car, the algorithm can hardly be blamed for reporting camera motion.
Only a human can declare which parts of the scene should be considered stationary.
When such a declaration is needed, it is made by computing offsets
for only a subregion (the sliver of prairie).

This deshaking approach emulates block-matching algorithms for motion estimation
(computing motion vectors, in the jargon of video compression),
which implicitly and approximately separates moving objects from a static background.\footnote{
Motion estimation is another example of using approximate background subtraction only as a means to an end,
in this case higher data compression.}
Indeed, this is a special case of deshaking video,
where camera shake is far more violent.  Elaborate video deshakers~\cite{deshaker,cinerella}
share features with our software, such as adjusting block size,
discarding low-contrast blocks, restricting analysis to a subregion, and border processing.

To quickly verify the deshaking,
the de-shaken images are converted to a video file.
While scrolling and flipping through the video,
if ostensibly static parts of the scene such as distant buildings jump around,
deshaking should be rerun.
This could be caused by the camera's actual motion exceeding the specified threshold for offsets,
or by too much motion in the subregion declared as stationary.

Having eliminated camera shake, we can now estimate the background.

\section{Background estimation}
\label{sec:bg}

For each image, we average a contiguous interval of its neighbors to
suppress any moving objects, leaving the nonmoving background.
Although skipping some neighbors (say, using only every tenth one~\cite{subsample}) speeds up calculation,
we avoid this shortcut because such subsampling leaves ugly gaps in motion trails (section~\ref{sec:trails}).

The width of the averaging window is manually adjusted to the image material, to the speed of foreground motion.
It typically ranges from 10 to 200 images.
It should be brief enough to track changing global illumination (common outdoors),
yet long enough to erase moving foregrounds.

The averaging over images is done per pixel.
Again, instead of averaging with a fast arithmetic mean,
it uses the geometric median~\cite{median}
to reject outliers, which in this case are
foreground objects found in a minority of the images.
For speed, these color calculations are done in RGB space.
Conversion to and from a perceptually smoother color space might reject more outliers,
but not enough to matter at this early stage in the signal chain.
Also, for this per-pixel task, the (now three-dimensional) geometric median is more suitable
than using color quantization to build a degenerate one-color palette~\cite{mediancut}.

Averaging can be conveniently implemented as a program that reads a
windowful of images to calculate one averaged image.
But running one program per image is inefficient:
an averaging width of 100 then forces each image file to be parsed 100~times.
Instead, a program that processes the entire sequence
parses each image only once (and then frees that memory when no longer needed).
This speedup lets a commodity 6-core CPU
combine a hundred 8~megapixel images (2.5~GB) in only 7~s.

\section{Foreground segmentation and boundary~refinement}
\label{sec:fg}

Each pair of original-background images is analyzed to estimate the foreground.
A rough estimate is found by noting each pixel position where
the colors differ by more than a threshold.
(Color difference is computed in YCbCr space with hue overweighted,
to encourage shadows, whose hue difference is slight, to remain background.)
This foreground-background segmentation is then refined by heuristics
progressing from small details to larger features.
Such binary classification followed by refinement is conventional.

As a first refinement, pinholes are removed:
any pixel classified oppositely from its four neighbors is flipped.\footnote{
This is like a simultaneous and symmetric {\tt clean} and {\tt fill} operation,
in the terminology of MATLAB's {\tt bwmorph} function.}
(Iterating this would remove more pinholes from pathological
checkerboards, but such rare cases are
better handled by larger-scale refinements.
Iterating until no pinholes remain would even risk non-termination.)

\subsection{Disk smoothing}
\label{sec:fgDisk}
The foreground-background boundary is smoothed further with a
disk-shaped structuring element.  For each pixel,
if the fraction of the pixel-centered disk that is foreground
exceeds a threshold, the pixel itself becomes foreground;
otherwise it becomes background.\footnote{
This generalizes the {\tt majority} operation of {\tt bwmorph}.}
This heuristic is repeated for gradually larger disks,
stopping before desirable details are smoothed away.
This fast approach can be thought of as a simplified
adaptive morphological operator~\cite{adaptive1,adaptive2}.

\subsection{Removing small objects}
Another heuristic exploits the known lower bound of an object's
area in pixels: pedestrians or cars can be only so small.
In this common case, connected component labeling (CCL)
efficiently identifies the objects and measures each one's area~\cite{rosenfeld,suzuki}.
Any object whose area is smaller than $10^{-4}$ times that of the whole image
is tossed back into the sea:
its pixels are relabeled as background.

Very thin objects are also culled.  Any object whose bounding box is improbably thin,
either absolutely or in the sense of aspect ratio, is relabeled as background.
This test is rarely worth generalizing from the axis-aligned case,
because such objects are usually sun breaking through clouds,
glinting suddenly on the long horizontal or vertical edges of buildings.

\subsection{Removing holes and near-holes}
Next, we remove holes (background regions) from foreground objects,
again with CCL.\footnote{
The assumption that objects lack holes is violated by
a quadruped whose legs cross when seen from the side,
and even by a biped whose legs and shadow encircle background.
But erroneously filling in such a hole turns out to be rather benign
for foreground overlaying (section~\ref{sec:trails}).}
The object and its bounding box are inverted and then labeled.
Each connected component is then a background region,
either a hole inside the object, or a region outside it.
We then relabel as foreground all components that do not touch the bounding box,
that is, all the holes.

Similarly, we remove near-holes, that is, background regions that
would be holes, were only a few background pixels reclassified as
foreground.  (If we call foreground `land' and background `sea,'
then holes are isolated lakes, and near-holes are rivers and bays.)
The object is dilated with an 8-connected structuring element.
If this expansion of foreground and shrinkage of background reduces any
bays to lakes, they will be detected by a fresh CCL and then removed.\footnote{
Preceding a CCL with a dilation has been
used for similar goals~\cite{almostconcom,lidar}.}
However, because of the dilation,
the holes being removed are one or two pixels too small.
Deleting the shrunken hole would still leave a narrow moat.
To remove the moat, we reapply disk smoothing
(section~\ref{sec:fgDisk}).
This entire dilate-delete-smooth operation is then iterated, keeping
track of the radius of the progressively larger smoothing disk,
again stopping before oversmoothing.
Finally, because near-hole removal very occasionally creates fresh small holes,
one more hole removal operation is performed.

\begin{figure}[htbp]
\centerline{\epsfig{figure=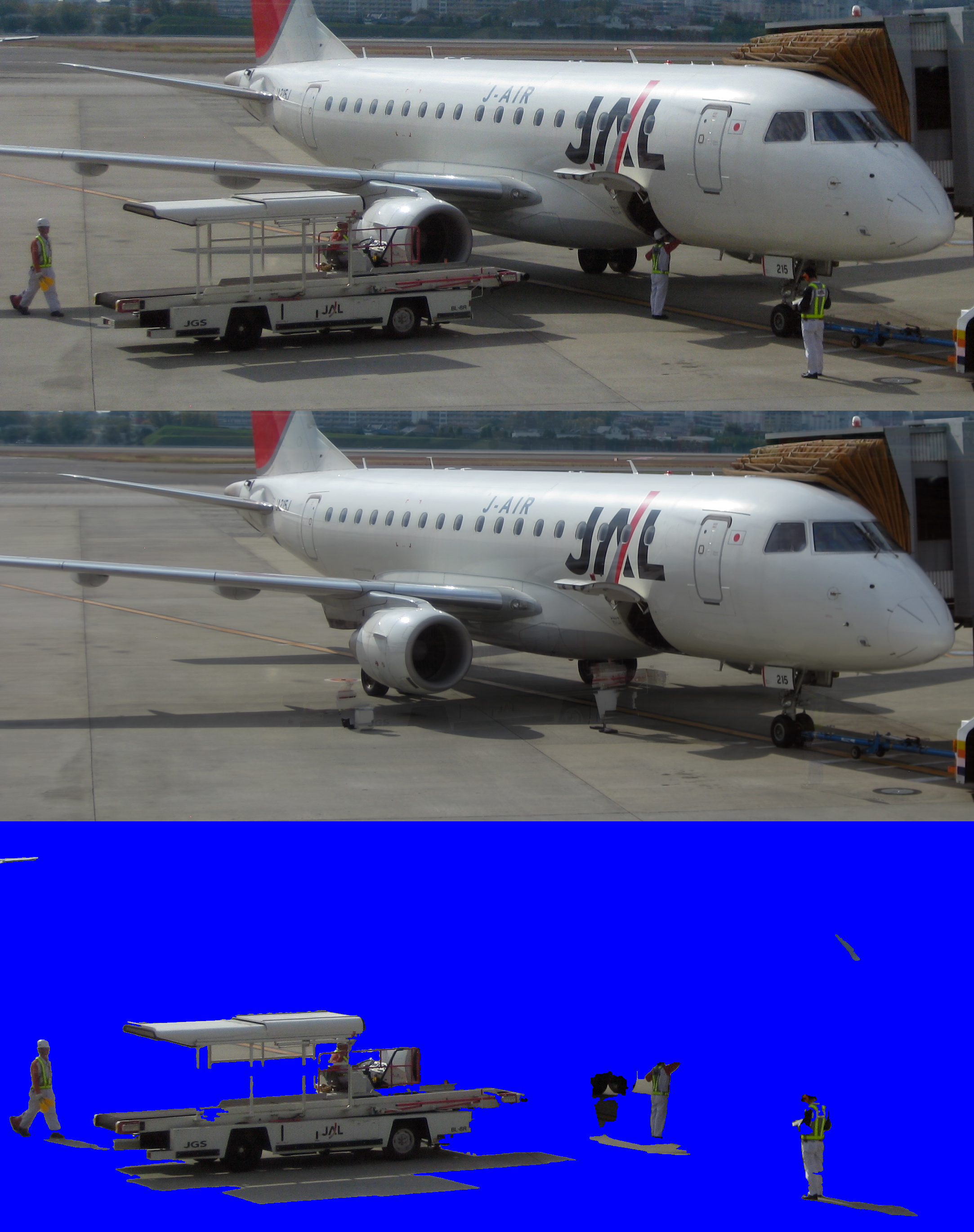, width=\linewidth}}
\centerline{\epsfig{figure=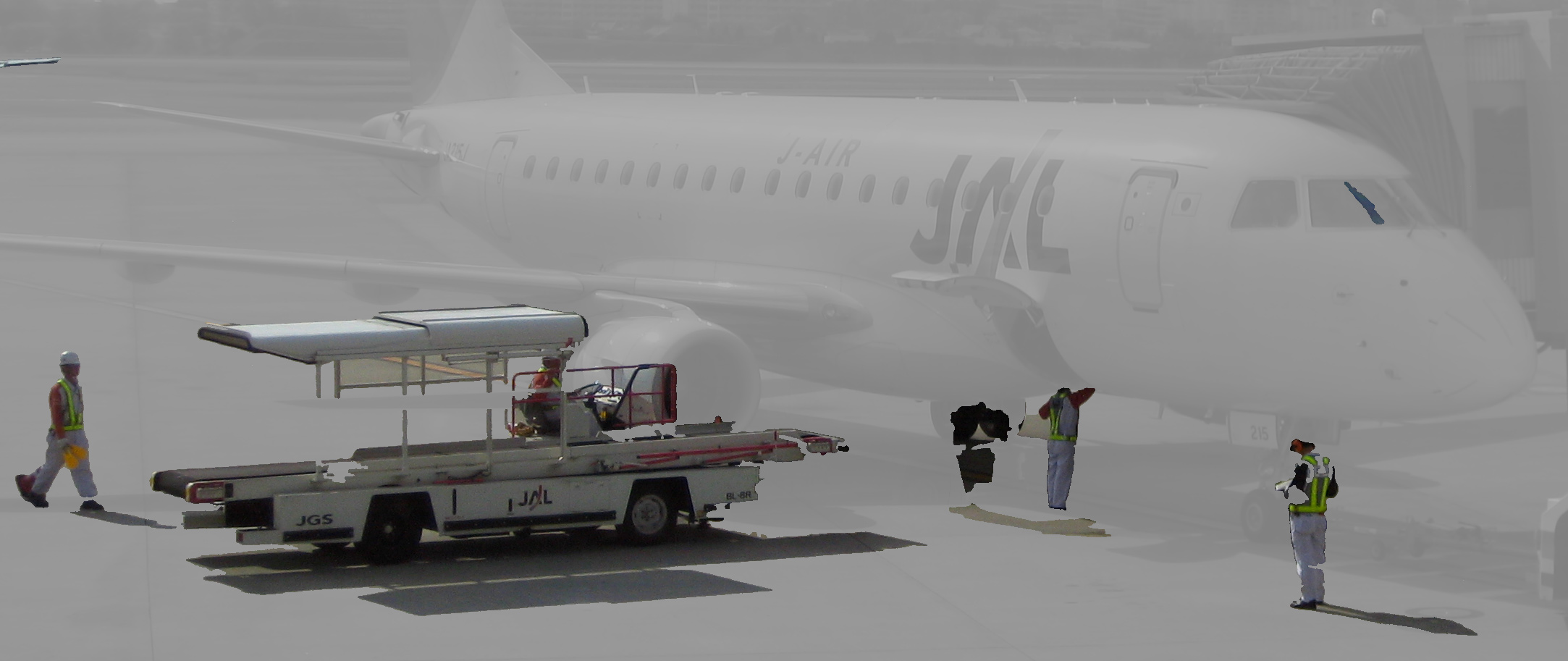, width=\linewidth}}
\caption{(i) Deshaken image; (ii) background averaged from this image's neighbors;
(iii)~extracted foreground;
(iv)~recombined foreground and background.}
\label{fig:fg}
\end{figure}

\section{Rendering motion trails}
\label{sec:trails}

Having split the original image stream into separate
background and foreground streams,
we can now recombine them in novel ways.
For example, if we desaturate or even erase the background,
then foreground objects gain prominence while losing context
(fig.~\ref{fig:fg}).

A more typical recombination, however,
overlays extra foregrounds onto each background.
For example, to increase the prominence of a momentary foreground event
(a vehicle moving so fast that it appears in only one image),
that foreground can be overlaid onto earlier and later backgrounds as well,
making that foreground appear a few seconds too early and disappear a few seconds too late.
Such instantaneous appearance and disappearance works well for infrequent events.
But if events are frequent enough to overlap in time, they should instead fade in and out.
During a fade, each pixel is a weighted sum of foreground and background pixels.
For a foreground from the $n$th frame, its weight starts at zero some time before $n$,
reaches unity at $n$, and then decreases again to zero.

\subsection{Pure fade-out}
When a scene contains many foreground objects,
or when each object is too small to let its pose indicate its direction of travel,
appearance is improved by omitting the fade-in.
A sharp onset clarifies both the object's direction of motion and its current position,
as distinct from its past positions.
Fade-ins, or fades that are asymmetric, reversed, nonmonotonic, or discontinuous,
are better suited to purely artistic goals.
(This also holds for audio processing.  Echos and reverberation improve the
realism of a dry recording, but a crescendo of reversed echos sounds unnatural and clever.
Non-causal processes are harder to interpret.)

\subsection{Multiple foregrounds}
Fading adds a complication:
multiple foregrounds may occupy the same pixel, either from multiple objects crossing paths,
or from one object moving so slowly that its successive positions overlap.
Then we must choose how to combine foregrounds, that is, how much to weight each one.
At a particular pixel, each foreground has its own intrinsic weight,
coming from the progress of its particular fade.
One choice is to rescale these foreground weights to sum to what was the heaviest one.
Another choice is to sum foregrounds, starting with the heaviest, until the accumulated weight reaches unity;
any weight still available is assigned to the background.

But in practice, blending multiple foregrounds looks confusing:
multiple objects occupying the same space are harder to interpret
than several copies of the same object spaced along a path.
So we simply choose the heaviest foreground---the most recent one, for pure fade-out---and ignore any others.
If its weight is less than unity, the remaining weight is assigned to the background.

\subsection{Length of motion trails}
Several constraints guide how long to make a trail.
It should be long enough to reveal transient events.
On the other hand, if the trail is too long, its object's opacity may vary
too slightly to maintain any sense of motion (top half of fig.~\ref{fig:nara}).
Also, long trails may overlap,
producing more clutter than clarity (fig.~\ref{fig:wrightgreen}).
Fortunately, evaluating different trail lengths is quick
because the overlaying is purely mechanical;
the expensive image analysis has been already completed.

When trails can indeed be made very long,
they look better with abrupt beginnings.
This can be done by replacing a linear fade-out
with one that is quadratic or even cubic.
This can be seen in the bottom half of fig.~\ref{fig:nara},
where both a deer and a family are much more pronounced than
their respective trails.

\section{Ghosts}
\label{sec:ghost}
For the simple task of adding motion trails,
we have implicitly defined foreground to be
whatever persists more briefly than the averaging window.
This dispenses with elaborate reasoning about
multiple occluding objects, but at the cost of producing an artifact.
When a formerly moving object holds still for longer than the averaging window,
the average starts to include that object:  the object becomes background.
Later, when it suddenly moves away,
the freshly revealed `true' background behind it---which had become
excluded from the average---is misclassified as foreground,
a ghost~\cite{ghost} of the true object, such as a car-shaped patch of pavement.

\begin{figure}[bp]
\centerline{\epsfig{figure=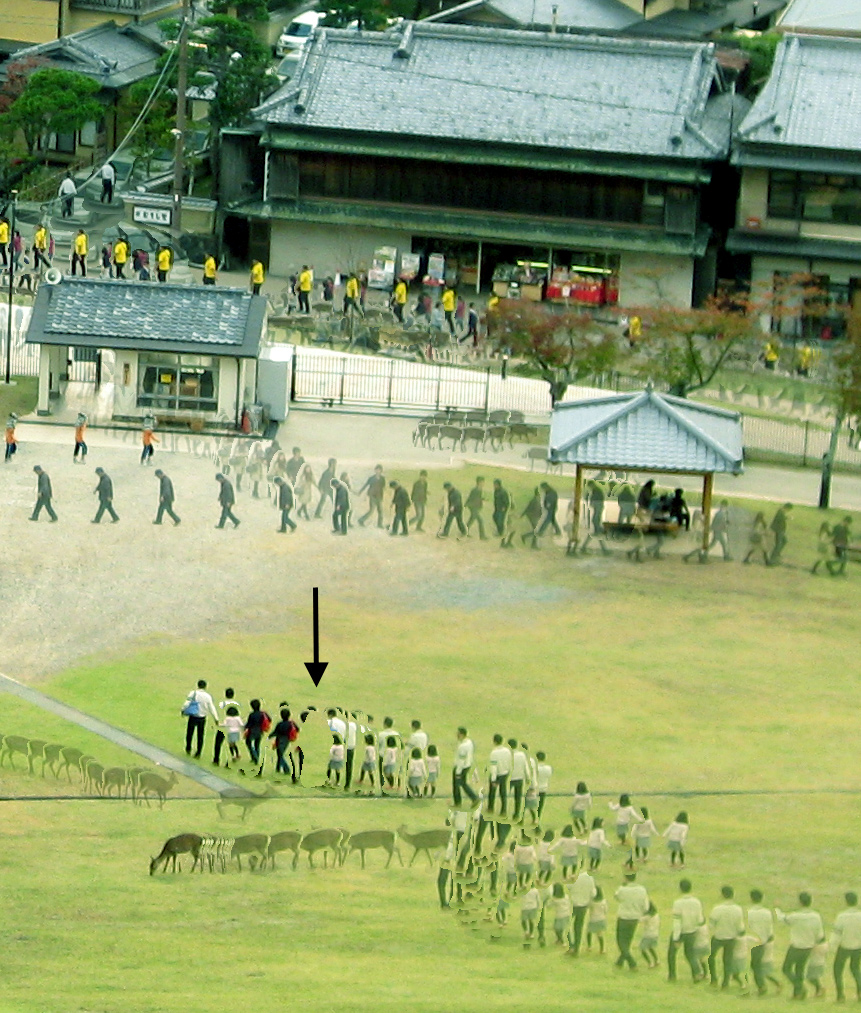, width=\linewidth}}
\caption{
A grassy ghost covering the motion trails of a family.
The mother had waited so long to photograph her husband and daughter
that she became classified as background.
When her position was suddenly vacated,
it was misclassified as fresh foreground.
Mount Wakakusa, Nara, 2012-11-04.}
\label{fig:nara}
\end{figure}

\begin{figure}[bp]
\vspace{4mm} 
\centerline{\epsfig{figure=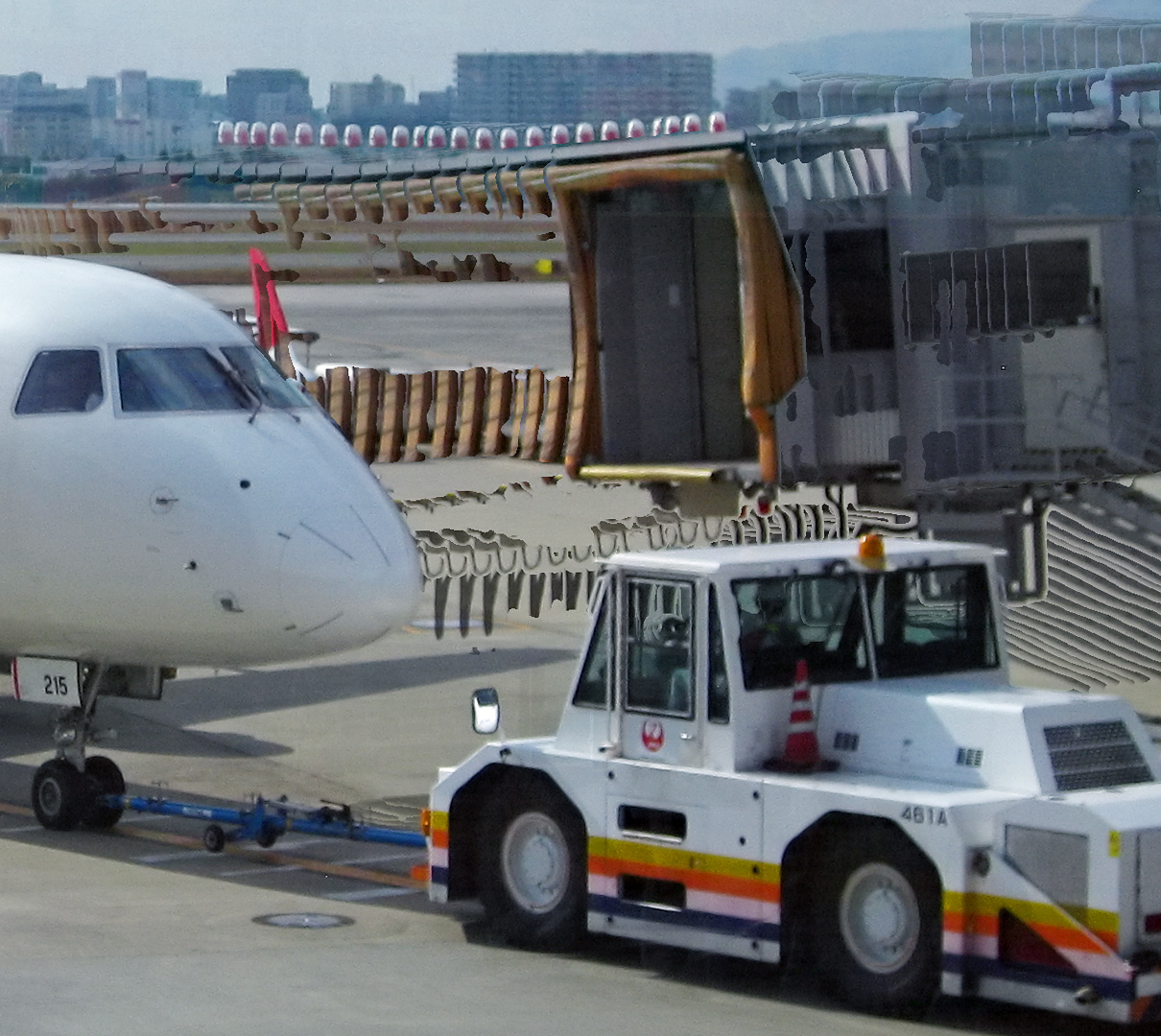, width=\linewidth}}
\caption{
Suddenly revealed pavement (a ghost) behind a retracting jet bridge,
misclassified as foreground and thus partly covering the jet bridge's light brown motion trail.}
\label{fig:ghost}
\end{figure}

In itself a ghost is harmless, because the misclassified region
exactly matches the background and is thus unseen.
But when the same or another object moves through that region,
the resulting motion trail may be covered by the ghostly outline of
a traffic pylon (fig.~\ref{fig:butterfly}),
a person (fig.~\ref{fig:nara}) or part of a jet bridge (fig.~\ref{fig:ghost}).

Ghosts can be suppressed by artificially increasing the foreground segmentation's
color difference threshold (section~\ref{sec:fg}),
but this greater tolerance for color variation also undesirably suppresses non-ghost foreground.
Ghosts can be eliminated outright by artificially widening the averaging window,
but this fails under varying illumination (section~\ref{sec:bg}).
Conversely, an extremely narrow window makes ghosts expire almost instantly,
but also creates many more ghosts.
In short, manipulating any averaging or segmentation parameters at all causes undesirable side effects.

\subsection{Detecting ghosts}
Instead of preventing ghosts in the first place,
we might try to detect and remove them after the fact.
An object may be a ghost if it is
near another object of similar compactness\footnote{
Compactness measures how much an object fills its bounding box or its convex hull.
It is a fast approximation of shape.} and area,
and its median color approximates that of its surroundings (itself subtracted from its dilation)
better than that of the other object.
This heuristic tries to distinguish an actual car from a car-shaped patch of pavement.
If the surroundings' color is too varied (not grass, pavement, or sky),
then the heuristic gives up and cannot accuse the object of being a ghost.

But even a heuristic this elaborate is easily fooled.
Two persons talking together for a while may overlap into a single large object.
When they walk away in different directions,
neither of the two new small objects match the size or shape of the old large one.
Thus, the heuristic fails to recognize the large ghost left behind.

Compactness-based heuristics fail even for a solitary object,
when the object's pose (and hence compactness) differs between rest and motion.
Common examples include walking contrasted with standing and sitting,
and moving vehicles contrasted with stopped ones whose doors or hatches are open.
Although it is tempting to relax the similarity thresholds for area, compactness, or color,
this reports so many false positives---mistakenly erases so much
foreground---that the reduction in occasional artifacts due to ghosts
is overwhelmed by the increase in inevitable artifacts due to missing foreground.

In short, robust removal of ghosts requires
object-level reasoning~\cite{subsample} such as
object tracking~\cite{track1,track2} or blob tracking~\cite{bramble}.
Such software may be too elaborate for merely adding motion trails.

\section{Conclusions}

Motion trails can be added to time-lapse video by
eliminating camera shake,
estimating the background by averaging,
isolating the foreground with several heuristics,
and then overlaying multiple foregrounds onto each background.
This novel combination is useful for fast-forward previewing.

Foreground segmentation would be improved by replacing the fixed threshold
for color difference with a per-pixel color covariance matrix~\cite{pfinder}.
The segmentation's robustness against background motion (curtains, trees, water)
and outdoor lighting changes could be further increased with
Bayesian classification~\cite{bayes}, a background mixture model~\cite{bmm},
or other elaborate schemes~\cite{piccardi}.

Finally, if an object-tracking stage is added to remove ghosts,
feedback between object tracking and foreground segmentation
would further improve the latter~\cite{track3}.

\bibliographystyle{IEEEtran}

\end{document}